\documentclass[10pt,twocolumn,letterpaper]{article}

\usepackage{iccv}              %

\definecolor{iccvblue}{rgb}{0.21,0.49,0.74}
\usepackage[pagebackref,breaklinks,colorlinks,allcolors=iccvblue]{hyperref}
\usepackage{amssymb} 
\def\paperID{41} %
\def\confName{ICCV}
\def\confYear{2025}

\title{Uncertainty-Aware ControlNet: Bridging Domain Gaps with Synthetic Image Generation}

\author{
  Joshua Niemeijer$^{1}$ \quad
  Jan Ehrhardt$^{2,3}$ \quad
  Heinz Handels$^{2,3}$ \quad
  Hristina Uzunova$^{3}$\\[1ex]
  $^{1}$German Aerospace Center (DLR) \quad
  $^{2}$University of Lübeck \\
  $^{3}$German Research Center for Artificial Intelligence (DFKI)\\
  \parbox{\textwidth}{
    \centering
    \small\ttfamily
    Joshua.Niemeijer@dlr.de,\quad \{jan.ehrhardt,heinz.handels\}@uni-luebeck.de,
    hristina.uzunova@dfki.de
  }
}
\newlength{\ilength}
\newcommand{\unacorn}{\textbf{Un{\large\color{pink}I}\hspace{-\ilength}{\small A}CorN}}

\newlength{\ilengthnb}
\newcommand{\unacornnb}{Un{\large\color{pink}I}\hspace{-\ilengthnb}{\small{A}}CorN}

\begin{document}
\maketitle
\begin{abstract}
Generative Models are a valuable tool for the controlled creation of high-quality image data. 
Controlled diffusion models like the ControlNet have allowed the creation of labeled distributions. Such synthetic datasets can augment the original training distribution when discriminative models, like semantic segmentation, are trained. 
However, this augmentation effect is limited since ControlNets tend to reproduce the original training distribution.

This work introduces a method to utilize data from unlabeled domains to train ControlNets by introducing the concept of uncertainty into the control mechanism. The uncertainty indicates that a given image was not part of the training distribution of a downstream task, e.g., segmentation. Thus, two types of control are engaged in the final network: an uncertainty control from an unlabeled dataset and a semantic control from the labeled dataset. 
The resulting ControlNet allows us to create annotated data with high uncertainty from the target domain, i.e., synthetic data from the unlabeled distribution with labels. In our scenario, we consider retinal OCTs, where typically high-quality Spectralis images are available with given ground truth segmentations, enabling the training of segmentation networks. 
The recent development in Home-OCT devices, however, yields retinal OCTs with lower quality and a large domain shift, such that out-of-the-pocket segmentation networks cannot be applied for this type of data. 
Synthesizing annotated images from the Home-OCT domain using the proposed approach closes this gap and leads to significantly improved segmentation results without adding any further supervision. 
The advantage of uncertainty-guidance becomes obvious when compared to style transfer: it enables arbitrary domain shifts without any strict learning of an image style. This is also demonstrated in a traffic scene experiment. 
The implementation is available:  \url{https://github.com/JNiemeijer/UnAICorN.git}
\end{abstract}

\section{Introduction}

\begin{figure}[t]
    \centering
    \includegraphics[width=\linewidth]{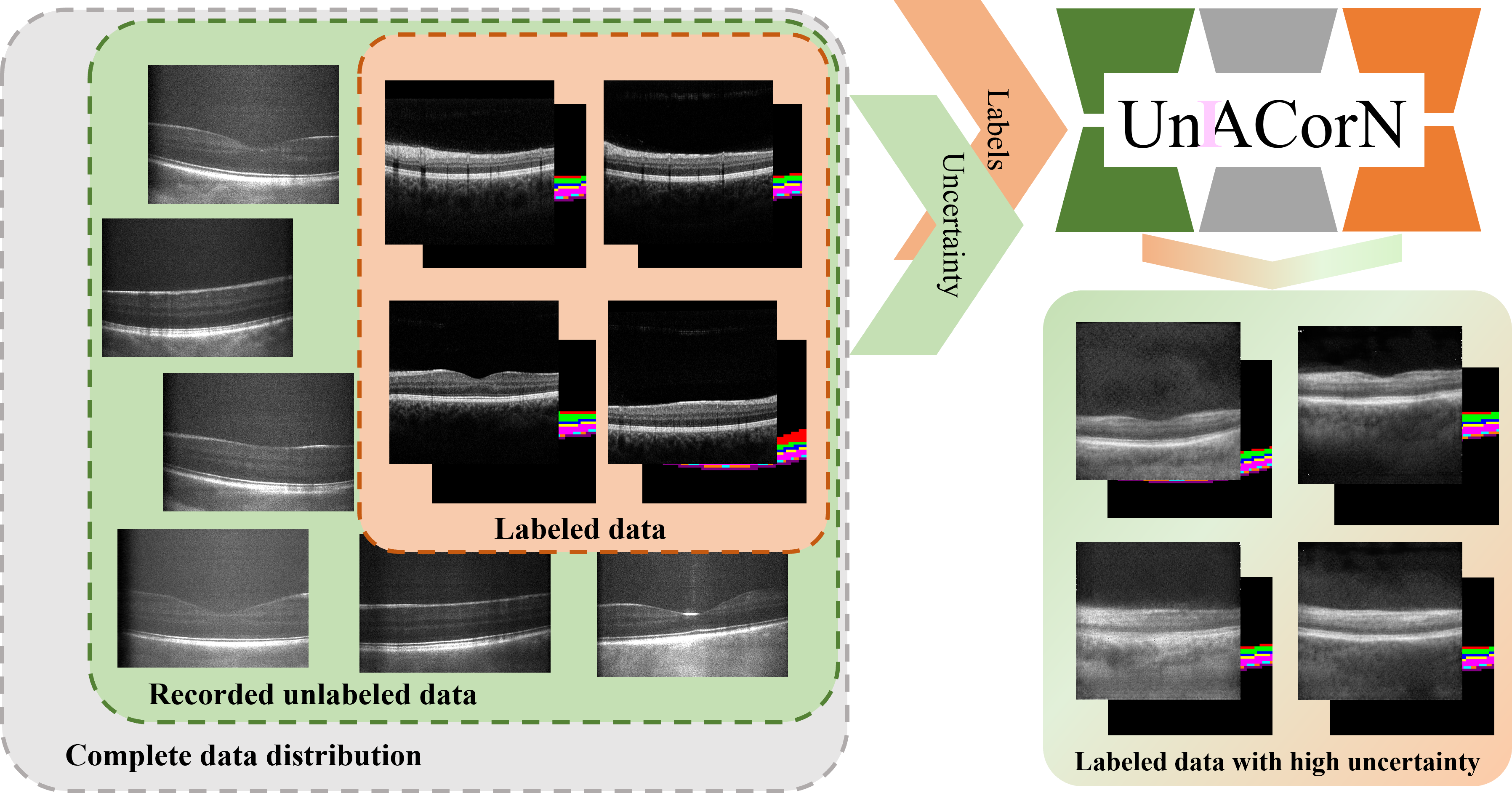} 
    \caption{We aim at learning the uncertainties of a recorded but unlabeled distribution and generating new labeled data containing high uncertainties.}
    \label{fig:dual-control-arch}
\end{figure}

With the shift of research interest towards deep neural network-based approaches, the problem of labeled data scarcity becomes increasingly prominent. Typically, discriminative models, like semantic segmentation networks, require large and representative training distributions with given ground truth annotations. 
However, such distributions are inherently hard to create. 
Firstly, manual labeling effort is costly and time-consuming. 
Secondly, the test data might expand to previously unknown points of the distribution, making the creation of a realistic representative training dataset impossible. 
This problem becomes even further aggravated when it comes to the medical image domain since data availability is generally limited, rare diseases lead to unexpected distributions, and privacy concerns need to be taken into account. 
Furthermore, most labeling tasks require expert knowledge about the specific domain. 
Thus, existing pre-trained methods are often ill-suited to newly acquired datasets (e.g., using a different device or a novel acquisition technique). 
A known technique to work against data scarcity, in general, is data augmentation. 
The most common data augmentation type includes simple image transformations like affine/elastic image deformations and adding noise to the image intensities. 
Such augmentations typically extend the training data distribution and, if chosen carefully, lead to better performance. 
However, with this simple type of augmentation, most cases would conform to the training data domain and would not perform well with large domain shifts in the test dataset. 
A more sophisticated way of data augmentation is enabled by generative methods, where new data domains can be synthetically produced. In order to enable the usage of the generated data for training, the data is commonly generated conditioned on given ground truth labels.  
Prominent examples of unpaired domain translation methods are CycleGANs \cite{cyclegan}, which can translate two unlabeled and unpaired distributions into each other and generate realistic images of a new domain. 
However, CycleGAN-generated images often do not conform to the shape and, thus, semantic labels of the source domain and cannot directly be applied for training. 
Ideas to overcome this problem are presented by approaches like CyCADA \cite{cycada}, or ACCUT \cite{seibel2024anatomical}.
However, these methods require a reliable segmentation approach for the target data domain, which might be challenging depending on the application.
Furthermore, there is a variety of unsupervised topology-preserving domain translation approaches specifically designed for medical images (e.g.,  \cite{uzunova2022synthesis,uzunova2020memory}). 

Naturally, led by the current state-of-the-art in image generation,  diffusion models that stand out for their exceptional capacity to produce realistic images and offer excellent control possibilities like ControlNet \cite{zhang2023adding,stracke2024ctrloralter,ye2023ip,lin2024ctrl}, approaches like MAISI~\cite{guo2024maisi} have emerged for large-scale image generation in the medical image domain. 
Even though ControlNets allow for the creation of high-quality labeled distributions by introducing image labels as a control mechanism, generative models like GANs and diffusion models are known to reproduce their training data distribution. 
This essentially limits their augmentation effect since unknown distribution parts that might be present in the test dataset will most likely not be synthesized. 
Instead of selective control of the generated image distribution, we want to shift the paradigm towards an automated control of the synthesis process in order to generate images that do not conform to the predetermined distribution of the generative model. 
However, we are regulated by explicit requirements of the downstream task.

We pursue the idea of generating labeled unseen image distributions, specifically targeting the relevance of the data for a downstream task. 
This requires a metric that enables a statement about data relevance. 
Inspired by methods of active learning \cite{mittal2025realistic}, where relevant images are selected for training based on epistemic uncertainty, we propose to utilize this metric in a generative manner. Previous works like TSynD~\cite{niemeijer2024tsynd}, where the generation process is steered towards data with high epistemic uncertainty, have shown the efficacy of this consideration. 
However, TSynD requires a time- and memory-consuming optimization of the latent space of a generative model and is only able to slightly adapt an input image to the given uncertainty level, offering neither a possibility to generate completely new data nor a guarantee of topology preservation of the input data.

In this work, we introduce \unacorn: an \textbf{un}certainty-\textbf{a}ware \textbf{Co}nt\textbf{r}ol\textbf{N}et, which enables the generation of labeled image data with high epistemic uncertainty towards a given discriminative model. The proposed method combines two independent conditioning strategies: 1) semantic conditioning on the image labels, trained on a given labeled data distribution, and 2) conditioning on the uncertainty, trained on an unlabeled image domain. In this way, unlabeled images can be utilized for training, as well, where their uncertainty for a particular segmentation model is mapped to image information, and a labeled data distribution with control over uncertainty levels can be produced. The main contributions of our work are:
\begin{itemize}
    \item We propose an \emph{uncertainty-based conditioning mechanism} for ControlNets, enabling the training on existing unlabeled datasets with large domain gaps without the need for any semantic control information.
    \item Unlike the conventional ControlNet, where only data from the labeled distribution can be generated, we enable the combination of a semantic label control and uncertainty control, leading to the synthesis of \emph{diverse labeled datasets}. 
    \item In contrast to traditional style-transfer methods, bound to the statistical distribution of the underlying data domains, our approach offers the possibility to \emph{sample data based on different uncertainty levels} and, thus, generate datasets covering larger parts of the distribution.
\end{itemize}

In our experiments, we evaluate \unacornnb\ on a distribution shift scenario in retinal OCTs, where we have a labeled distribution of the commonly acquired Spectralis OCTs and an unlabeled distribution of the novel self-examination technique HOME-OCTs \cite{von2022self}, that are structurally different from each other (examples in Fig.~\ref{fig:example_imgs}) \cite{seibel2024reducing,seibel2024anatomical,niemeijer2023overcoming}. 
Hence, we learn control over the class on the Spectralis dataset and control over uncertainty on both. 
In our experiments, we demonstrate that our approach to data generation allows for improved training when compared to common style transfer approaches.
However, \unacornnb\ is not yet another domain translation method: its ability to generate data based on uncertainty enables the representation of arbitrary domain shifts without learning a particular image style, which we illustratively show for traffic scene images.

\section{Methods}
\label{sec:method}
\begin{figure}[tb]
    \centering
    \includegraphics[width=\linewidth]{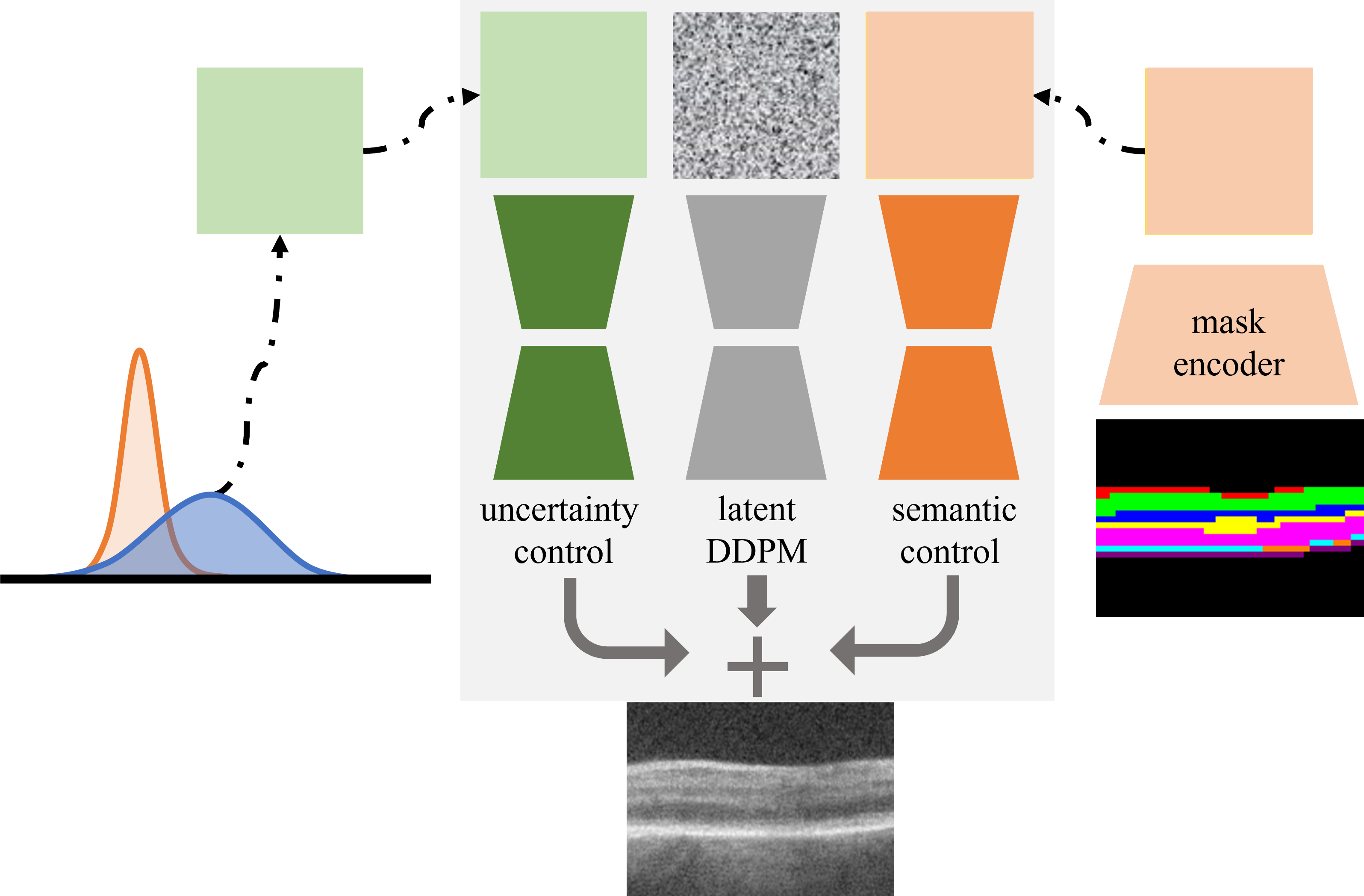}
    \caption{During inference,  we infer both the   Semantic-ControlNet and the Uncertainty-ControlNet in each diffusion step. 
    Finally, a weighted sum of both is computed, allowing for control over the segmentation mask and the uncertainty in the image.  }
    \label{fig:dual-control-arch}
\end{figure}
In this work, we present an "Uncertainty-Aware ControlNet":\unacornnb.
Our goal is to create training data distributions for a discriminative model, in this case, a segmentation neural network. 
The distributions should have the following properties: the data should match given ground-truth labels (``labeled''), and they should be highly tailored to the particular task (``relevant''). 

{\color{black}In order to fulfill the first property, i.e., ``labeled'', we utilize the ControlNet \cite{zhang2023adding} approach on a labeled dataset  coming from a distribution $(x,y)\sim p(\mathcal{X^L},\mathcal{Y})$, with $\mathcal{X^L}$ denoting images acquired with specific devices or parameters and $\mathcal{Y}$ are label maps of the underlying anatomy or scene. 
The resulting Semantic-ControlNet can create images $\hat{x}$ that correspond to a given label map. 
However, the generated images will correspond to the labeled data distribution $\hat{x}\sim p(\mathcal{X^L})$ (see Fig. \ref{fig:semantic-control}). 
Training on such data, therefore, does not provide additional information to the labeled data.}

{\color{black}To fulfill the second property, i.e., ``relevant'', we need to condition the generated images to a certain appearance.  
A common approach is to use active learning to select data that is particularly relevant to a task and requires labeling. 
So-called acquisition functions are utilized to select images with a high epistemic uncertainty for manual labeling. 
The new labeled data can then serve as additional input to a ControlNet.  
In our approach, we utilize this idea in an alternative way, which does not require human intervention for labeling. 
We propose to employ the (epistemic-) uncertainty of a downstream discriminative model as an indicator function for identifying relevant parts of the unlabeled distribution.
Therefore, in our application, we train a second separate ControlNet conditioned on the epistemic uncertainty of a segmentation model trained on the labeled distribution. 
Since we can determine the uncertainty in an unsupervised way, we can train this Uncertainty-ControlNet on unlabeled but recorded (known) 
distributions $x\sim p(\mathcal{X^U})$.}
The resulting Uncertainty-ControlNet should, therefore, be able to create ``known-unknown'' information, i.e., image information that was recorded but is not labeled and is difficult for the segmentation network to process.

Given the two models, the Semantic-ControlNet and the Uncertainty-ControlNet, we have control over the two desired properties: 1. labels, and 2. relevance. 
In order to create the image, we fuse the predictions in each diffusion step as shown in Fig. \ref{fig:dual-control-arch}. 

\subsection{Pre-Training}
\begin{figure}[tb]
    \centering
    \includegraphics[width=\linewidth]{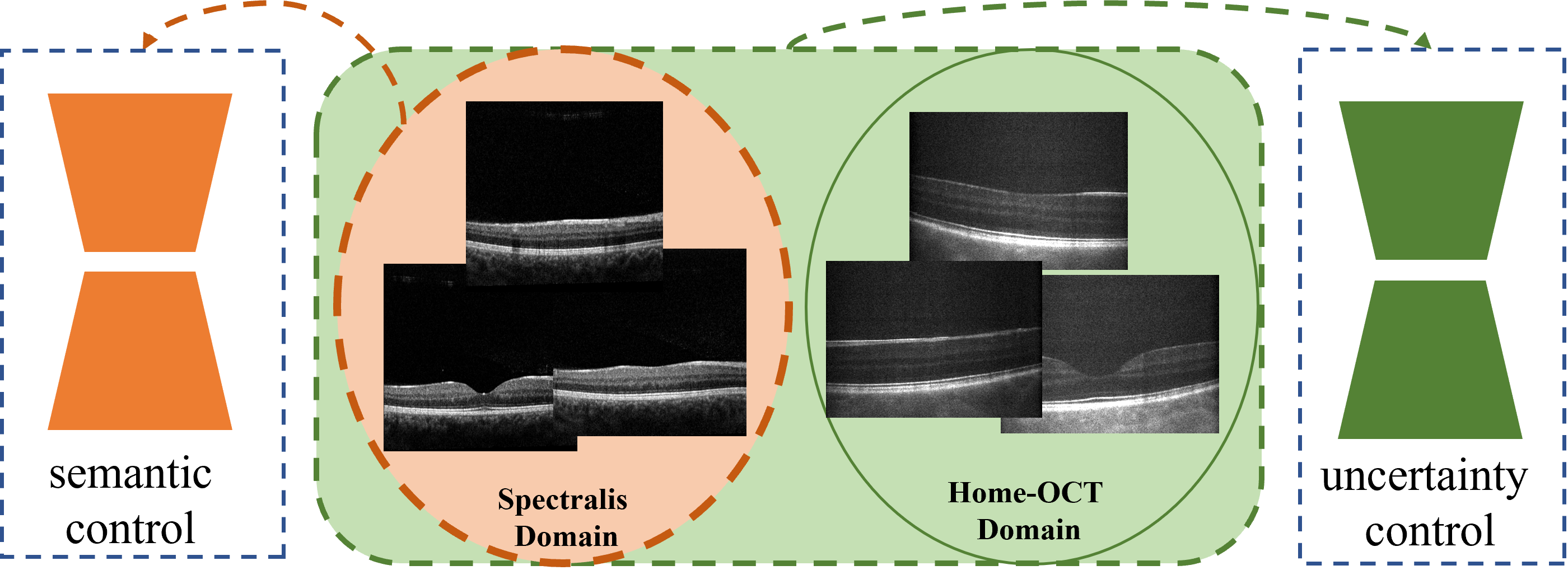} 
    \caption{The Semantic-ControlNet is trained on the labeled Spectralis Domain. 
    The Uncertainty-ControlNet is trained on both the Spectralis and Home-OCT domains.}
    \label{fig:pre-train}
\end{figure}
Our approach consists of three building blocks: a denoising diffusion probabilistic model (DDPM), the Semantic-ControlNet, and the Uncertainty-ControlNet.

\textbf{DDPM pre-training:}
We assume/ train a DDPM with high expressiveness to create a wide variety of images, as is common for current foundation models. 
In particular, the DDPM must be able to generate images from both distributions $\mathcal{X^L}$ and $\mathcal{X^U}$, i.e., notably images that were not used during the training of the downstream task. 
Pre-training or fine-tuning of the DDPM is therefore only necessary if the use case involves new acquisition devices or rare imaging conditions, and can then be performed using unlabeled data.
Such a case is given for our medical use-case; for the automotive use-case, we can re-utilize existing foundation models.  
In our approach, we employ latent diffusion models (LDMs)~\cite{rombach2022high}, i.e., all images are transformed into a low-dimensional latent space representation $z\in\mathcal{Z}$ by a pre-trained encoder-decoder model. 
The DDPM predicts the noise in each denoising time-step $t$ :
\begin{equation}
\epsilon(z_t,t) = \mathcal{LDM}(z_t,t),
\end{equation}
where $z_t$ is a noisy version of the encoded latent space representation $z=\mathcal{E}(x)$ of image $x$,
where $\mathcal{E}$ denotes the encoder.

\begin{figure}[tb]
    \centering
    \includegraphics[width=\linewidth]{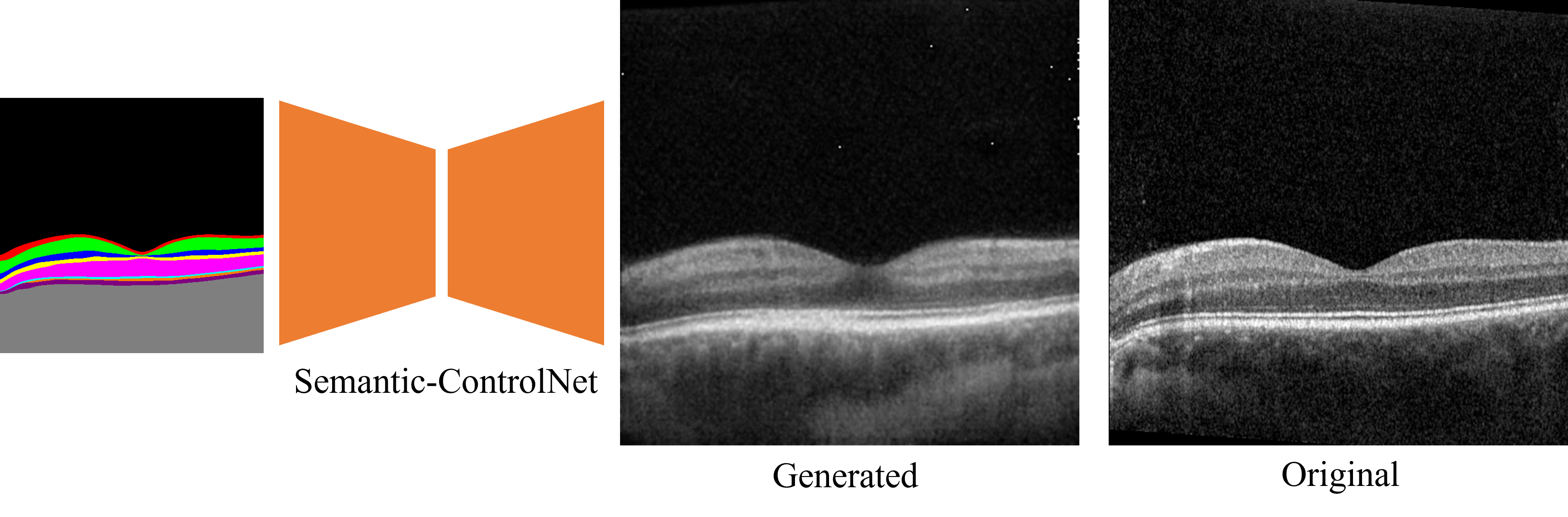} %
    \caption{The images generated by the Semantic-ControlNet match the distribution of the labeled images.}
    \label{fig:semantic-control}
\end{figure}
\textbf{Semantic-ControlNet:}
{
As shown in Fig.~\ref{fig:pre-train}, the Semantic-ControlNet is trained on the labeled data $(x,y)\sim p(\mathcal{X^L}, \mathcal{Y})$ only. }
As Fig. \ref{fig:semantic-control} shows, it, therefore, reproduces the labeled distribution during inference.
We denote the semantic control network as \(\mathcal{C}_S\), which takes as input the current latent space $z_t$ and the time step \(t\) as well as an encoded control image \(c_{s}=\mathcal{E_S}(y)\) representing the segmentation mask. The network \(\mathcal{C}_S\) generates a noise prediction:
\begin{equation}
\epsilon_S(z_t,c_s,t) = \mathcal{C}_S(z_t,c_s,t).
\end{equation}
The encoder $\mathcal{E_S}$ is trained along with the ControlNet. Its task is to adapt the spatial size of the label map to the size of the latent space without introducing artifacts, as would occur with plain down-sampling.

\textbf{Uncertainty-ControlNet:}
A segmentation network $\hat{y}=\mathcal{S}(x)$ was pre-trained based on labeled data $(x,y)\sim p(\mathcal{X^L},\mathcal{Y})$.  We utilize the uncertainty of the pre-trained segmentation network $\mathcal{S}$ as a condition for the training of the Uncertainty-ControlNet. As shown in Fig.~\ref{fig:pre-train}, uncertainty measurements of both image distributions, labeled and unlabeled images, are used for training. 
Our measure of uncertainty is the pixelwise entropy:
\begin{equation}
H(p(y|x_{i,j})) = -\sum_{y\in\mathcal{Y}} \hat{p}(y | x_{i,j}) \log(\hat{p}(y | x_{i,j})).
\end{equation}
The pixelwise entropy itself contains a lot of information about the spatial structure of the image. 
In our use case, this is not a desirable feature, as the Semantic-ControlNet generates the image structure, while the uncertainty describes the confidence of the segmentation network with the image content.
We, therefore, compute the normalized mean uncertainty of the image and scale it to the range $[0,100]$:   
\begin{equation}
    \mathbf{U}_H(x) = \frac{100}{N\cdot log(L)} \sum_{i,j} H(p(y|x_{i,j})),
    \label{eq:meanunc}
\end{equation}
where  $N$ is the number of image pixels, and $L$ is the number of labels in the segmentation mask.

\begin{figure}[tb]
    \centering
    \includegraphics[width=\linewidth]{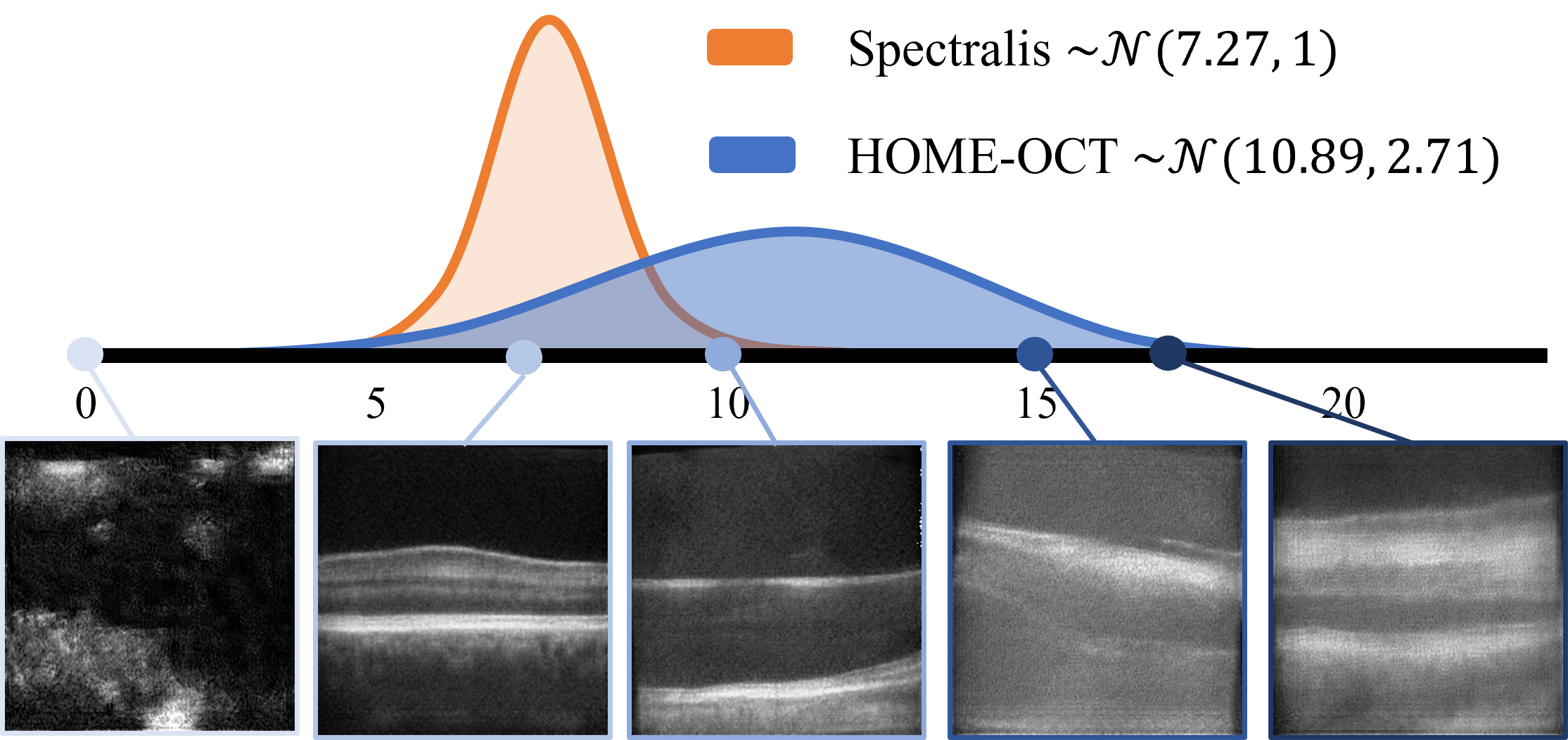} 
    \caption{The normal distributions of the average image uncertainties of the segmentation network applied to the Spectralis and Home OCT data. 
    We sample from these distributions and feed the resulting uncertainty to the Uncertainty-ControlNet. 
    The response to different uncertainties can be seen in the bottom row of images. }
    \label{fig:sample}
\end{figure}

The resulting uncertainty value is expanded to a control image where each pixel has the value of the average uncertainty.
This control image serves as input to the Uncertainty-ControlNet.
During inference, we can then sample from the uncertainty distributions. 
Fig.~\ref{fig:sample} shows the different distributions of average image uncertainty of the labeled and unlabeled datasets. 
As visualized, the images generated corresponding to given uncertainty values shift from the style of the labeled distribution to the style of the unlabeled distribution with increasing uncertainty. 
We denote the  Uncertainty-ControlNet as \(\mathcal{C}_U\). 
The \(\mathcal{C}_U\) network takes a latent representation \(z_t\), the control image \(c_{u}\) representing the average uncertainty, and a time step \(t\) as inputs, and produces a noise prediction:
\begin{equation}
\epsilon_S(z_t,c_u,t) = \mathcal{C}_S(z_t,c_u,t).
\end{equation}

\subsection{Inference}
Fig. \ref{fig:dual-control-arch} shows the final architecture that is used during inference.
We utilize the segmentation labels from the labeled distribution as input for the Semantic-ControlNet. 
At the same time, we infer the Uncertainty-ControlNet. 

\textbf{Sampling:}
The uncertainties are sampled from a Gaussian distribution corresponding to the uncertainty distribution of the unlabeled data. 
To create this Gaussian, we first apply our segmentation network to the images from the unlabeled distribution and determine the mean and standard deviation.
During sampling, we then randomly sample from the resulting Gaussian and create the uncertainty-control image from it. 
Therefore, the Uncertainty-ControlNet is steering the generation process to create images from the unseen distribution in each diffusion step. 

\textbf{Weighting:}
During the parallel inference of the Semantic-ControlNet and the Uncertainty-ControlNet, we fuse the down and mid-block noise prediction by computing a weighted sum. 
Let \(\epsilon_U(z_t,c_u,t)\) and \(\epsilon_S(z_t,c_s,t)\) denote the noise predictions from the Uncertainty-ControlNet and the Semantic-ControlNet, respectively, for an input \(x\) at time step \(t\). The fused noise prediction is given by
\begin{equation}
\epsilon_{\text{fused}}(z_t,c_u,c_s,t) = \alpha\, \epsilon_U(z_t,c_u,t) +\, \epsilon_S(z_t,c_s,t)
\end{equation}
where \(\alpha \in [0,1]\) controls the contribution of the Uncertainty-ControlNet.
$\alpha$ is empirically determined.
The LDM and the fused noise prediction of the ControlNets define the final noise prediction:
\begin{equation}
\epsilon(z_t,t,c_s,c_u) = \epsilon_{\text{fused}}(z_t,c_s,c_u,t) \oplus \mathcal{LDM}(z_t,t),
\end{equation}
where $\oplus$ is a $1\times1$ convolution. 

\subsection{Assembling the Bricks}
\begin{figure}[tb]
    \centering
    \includegraphics[width=\linewidth]{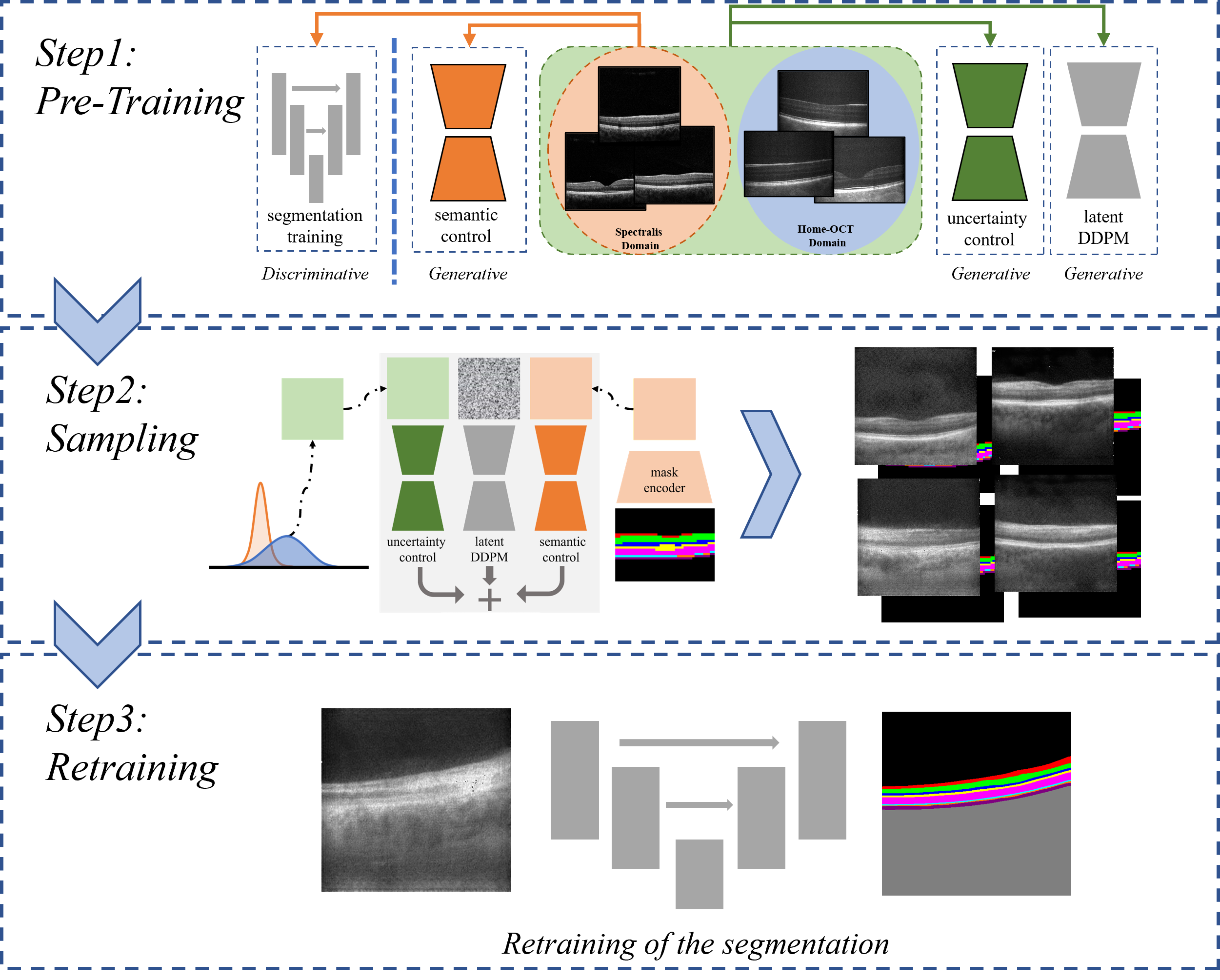} 
    \caption{Schematic depiction of the three steps of our approach. Step 1: The generative blocks and the segmentation model are trained. Step 2: The new training distribution is sampled. Step 3: The segmentation network is retrained. }
    \label{fig:steps}
\end{figure}

The final architecture of the \unacornnb~approach consists of three main steps, as shown in Fig.~\ref{fig:steps}.
First, we pre-train the diffusion blocks and the segmentation model. 
The diffusion blocks are the backbone LDM, the Semantic-ControlNet, and the Uncertainty-ControlNet.
During the training of the latter, the frozen pre-trained segmentation model on the labeled images provides the uncertainty control for all images.

The second step is the data generation using the pre-trained generative blocks as shown in Fig.~\ref{fig:dual-control-arch}. For this, uncertainty values from a normal distribution parametrized with the mean and standard deviation based on the real Home-OCT images are sampled and input to the Uncertainty-ControlNet, while a label map from a Spectralis image is used as input to the Semantic-ControlNet. This way, a new labeled dataset with high uncertainties is generated.

Finally, in the third step,  we retrain our discriminative network on the generated data. This process reduces the uncertainties from the model and yields improved segmentation results for out-of-distribution data. 

\section{Experiments and Results}
In our experiments, we mainly pursue a medical image processing scenario based on retinal OCTs. 
Here, we consider two structurally different distributions of  OCT-data coming from different sensor devices: Spectralis and HOME-OCT, generally assuming the Spectralis to represent the labeled data distribution and HOME-OCT to be unlabeled. We show that we can create a labeled distribution based on the segmentations from the Spectralis domain, containing the uncertainties observed in the Home-OCT domain, and compare the results to style transfer approaches. 
Additionally, we show an application to traffic scenes to prove the generalization ability of our method.

\subsection{Data}
\noindent \textbf{Retinal OCTs:}
In our experiments, an in-house OCT dataset from 50 healthy subjects is used \cite{seibel_oct}.
For each subject, both eyes are recorded with the Spectralis OCT device (Heidelberg Engineering, Heidelberg, Germany) and a prototype of a self-examination low-cost full-field (HOME) OCT device, yielding two different  OCT image domains (Fig. \ref{fig:example_imgs}). 
As typical for retinal OCTs, 2D B-Scans are extracted from the images and resampled to 256$\times$256 pixels, resulting in $\sim 10000$ images per domain. Both domains contain ground-truth segmentations of the anatomical retinal layers; however, the HOME-OCT labels are only considered for evaluation in our experiments. \\ 

\noindent \textbf{Street Scenes:}
The Cityscapes~\cite{Cordts2016} dataset (DCS) contains images taken from German cities under a stable sensor setup in very consistent environmental conditions. 
The ACDC~\cite{Sakaridis2021} dataset is taken in Zürich in different weather conditions. 
It contains rainy, snowy, foggy, and night scenes. 
Since the DCS only contains day images with few lighting/weather conditions, there is a large domain gap between the datasets. 
Both of these datasets provide semantic segmentation annotations. 
We utilize DCS as our labeled distribution and ACDC as our unlabeled distribution.

\subsection{Uncertainty-guided Retinal OCT Generation}
The primary objective of this experiment is to synthesize data that induces high uncertainty in a given pre-trained retinal layer segmentation model. 
The synthetic data is leveraged to refine the model through retraining, enhancing segmentation performance on previously unseen image domains.
In our experiments, we consider two image domains: a labeled source domain (Spectralis) and an unlabeled target domain (HOME-OCT). 

\subsubsection{Models}

Our method consists of three main building blocks: 
The backbone DDPM, the Semantic-ControlNet, and the Uncertainty-ControlNet. 

\textbf{Backbone DDPM:} 
For the synthesis of data, we assume that a generative model pre-trained on multiple domains exists. Due to the specificity of the retinal OCT data, here, a latent diffusion model based on~\cite{LDPMBackbone} was trained from scratch with the given datasets. However, with the current shift of research focus towards foundation models, this step may easily be omitted in the future. We plan on releasing the trained diffusion model.

\textbf{Semantic-ControlNet:}
We train the Semantic-ControlNet on the Spectralis domain for $150$ epochs with a learning rate of $2.5\text{e-}5$. 

\textbf{Uncertainty-ControlNet:}
The Uncertainty-ControlNet is trained on the combined unlabeled and labeled distributions. 
As described in section~\ref{sec:method},  we use the inference of a pre-trained segmentation network to compute the average entropy over the pixels according to Eq.~\ref{eq:meanunc}. 
The average entropy is assigned to every pixel of the control image that serves as input to the Uncertainty-ControlNet. 
This training is done for $150$ epochs with a learning rate of $2.5\text{e-}5$.

 Furthermore, for the uncertainty calculation, a retinal segmentation network is required. 
 Here, we train a U-Net-architecture \cite{ronneberger2015u} with Dice-loss for a maximum of 400 epochs using early stopping and a learning rate of $1\text{e-}4$. For uncertainty estimations, we pre-train the retinal layer segmentation neural network on the Spectralis data only and use the network's inference to calculate the uncertainty on both image domains.

\subsubsection{Data Generation}
During inference, we infer the diffusion process with $1000$ time-steps. 
In each time step, the noise prediction of the Semantic-ControlNet and the Uncertainty-ControlNet are fused by computing a weighted sum. 
We empirically found the weight factor $\alpha=0.4$ for the noise prediction of the Uncertainty-ControlNet to work well. 
For each of the training images of the labeled distribution, we create a synthetic image. 
Generally speaking, our method would allow for the creation of larger distributions. However, to ensure fairness in comparison with the style transfer approaches used here, we create the same number of images. 
To introduce more diversity into the distribution, we sample from the distribution of average image uncertainties from the unlabeled domain. 
This distribution is modeled as a Gaussian $\mathcal{N}(10.89, 2.71)$ as shown in Fig.~\ref{fig:sample}, where the mean and standard deviation are determined for the unlabeled data. 
Note that higher uncertainty values correspond to images that would be more challenging for segmentation (noisier, less visible layers), whereas lower uncertainty values, closer to the mean of the Spectralis distribution, yield images similar to the real Spectralis data.

\begin{figure}[tb]
    \centering
    \includegraphics[width=\linewidth]{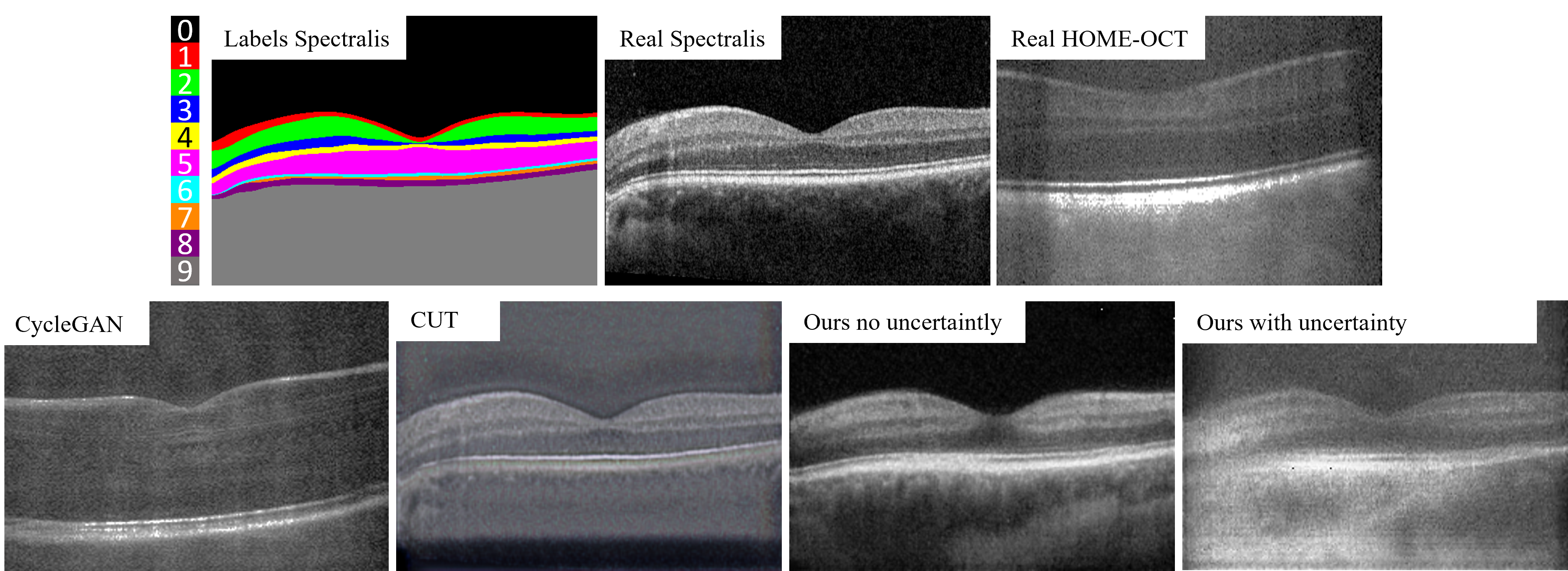} %
    \caption{
    Example image data of the real Spectralis and HOME-OCT images (top row); and synthesized images (bottom row), from left to right: generated using CycleGAN; generated using CUT; generated using only the Semantic-ControlNet; generated using the semantically- and uncertainty-conditioned ControlNets.}
    \label{fig:example_imgs}
\end{figure}

\subsubsection{Results}
As a surrogate of data quality in our experimental setup, we use generated and real datasets to train a retinal layer segmentation U-Net and observe its segmentation accuracy behavior. For comparison, we chose two common generative approaches: cycleGAN \cite{cyclegan} and CUT \cite{cut}, which are trained to generate domain-translated data (Spectralis$\rightarrow$HOME-OCT) in an unsupervised fashion. We generate training datasets of a rigid size ($\sim 1100$ images) using the different approaches and evaluate on a rigid size test set ($\sim 1100$ images) consisting of real HOME-OCTs. Overall, the following training setups are evaluated: real HOME-OCT images (upper bound), real Spectralis OCTs (lower bound); synthetic images generated by CylceGAN, CUT, and the proposed \unacornnb; synthetic image datasets combined with real Spectralis data. 
The results of this experiment can be seen in Tab.~\ref{tab:results}, where we show the mean intersection over union values (mIoU, also known as Jaccard index) over all retinal labels using the different training settings. Overall, the best training scenario features the dataset consisting of our images and the real Spectralis data, demonstrating the efficiency of our approach for data augmentation. 
Using the uncertainty-guided dataset generation, we achieve a mIoU of 0.65, which is a vast improvement compared to training on the real Spectralis data with a large domain gap. %

The results also show that the CycleGAN-generated training dataset is the worst-performing in this use case. 
This is because CycleGAN does not offer any semantic consistency mechanism, leading to realistic synthetic HOME-OCT images that do not align with the shape and labels of the corresponding Spectralis images. 
Thus, an inaccurately labeled data distribution is generated (example in Fig.~\ref{fig:example_imgs}). 
CUT, on the other hand, offers unsupervised semantic consistency and results in label-conform data distributions, as evidenced by better segmentation results. 
However, as visible in Fig.~\ref{fig:example_imgs}, the established domain shift is not as strong as the one achieved by our method. 

\addtolength{\tabcolsep}{-0.4em}
\begin{table}[]
    \centering
    \small
    \begin{tabular}{l|cc|ccc|ccc}
        &\multicolumn{2}{|c|}{real data}& \multicolumn{6}{c}{synthetic data}\\
        &
        HOME &
        Spect&
        Cycle&
        CUT &
        ours&
        Cycle& 
        CUT&
       ours \\
        lbl&\tiny{upper bound}&\tiny{lower bound}&&&&\multicolumn{3}{|c}{+real Spectralis}\\\hline
        0& 0.99&0.86&093&0.85&0.83&0.92&0.90&\textbf{0.96}\\
        1&0.71&0.33&0.45&0.41&0.35&0.49&\textbf{0.53}&0.51\\
        2&0.86&0.51&0.65&0.58&0.60&0.75&0.73&\textbf{0.78}\\
        3&0.69&0.37&0.47&0.52&0.41&0.59&0.33&\textbf{0.64}\\
        4&0.61&0.41&0.35&0.40&0.21&0.50&0.50&\textbf{0.60}\\
        5&0.86&0.65&0.41&0.63&0.65&0.60&0.74&\textbf{0.82}\\
        6&0.67&0.28&0.13&0.17&0.37&0.19&0.36&\textbf{0.40}\\
        7&0.60&0.36&0.01&0.25&\textbf{0.44}&0.31&0.38&0.35\\
        8&0.74&0.61&0.10&0.51&\textbf{0.69}&0.58&0.62&0.60\\
        9&0.96&0.84&0.48&0.80&0.77&0.74&0.81&\textbf{0.87}\\\hline
        $\varnothing$ all&0.77&0.52&0.40&0.51&0.53&0.57&0.59&\textbf{0.65}\\
        $\varnothing [1-8]$&0.72&0.44&0.32&0.43&0.46&0.50&0.52&\textbf{0.59}
        
    \end{tabular}
    \caption{
    Mean Jaccard index values of the retinal segmentation network, trained on different datasets: real HOME-OCT images (upper bound), real Spectralis OCTs (lower bound); synthetic images generated by CylceGAN, CUT, and the proposed \unacornnb; synthetic image datasets combined with real Spectralis data. 
    The last two rows show mean values over all labels and labels $[1-8]$, excluding the background labels.}
    \label{tab:results}
\end{table}

To evaluate the similarity between the domains of the generated data and the real HOME-OCT data, we compute the commonly used FID (Fréchet inception distance) metric, which measures the similarity between distributions (Tab.~\ref{tab:FID}). These results emphasize that CycleGAN indeed establishes a successful domain adaptation, as it is, per design, a style-transfer approach. However, due to the large spatial bias in the Spectralis and HOME-OCT domains, CycleGAN is simply able to map the distribution of the HOME-OCT data containing very large spatial shifts w.r.t. the given retinal labels. 
Furthermore, the large FID values of the semantically constrained methods, CUT and \unacornnb, show that the FID score is likely influenced by the spatial bias as opposed to only the image style of the distributions. Apart from that, the FID score shows that \unacornnb~is able to slightly reduce the gap between the domains by achieving $FID=132.5$ for the generated data, compared to $FID=152.6$ for the real Spectralis data. 
Also, the FID value between the real Spectralis and real HOME-OCT data distributions is very similar to the FID values of the generated data and the real Spectralis ($152.6$ vs. $151.9$), again suggesting a good domain adaptation ability. Additionally, we measured the mean uncertainty values over the generated and real distributions as suggested in Eq.~\ref{eq:meanunc}. The proposed \unacornnb~approach, as expected, yields a high mean epistemic uncertainty value of $12.85$, which is significantly higher than the real Spectralis data ($7.27$) and similar to the mean over the real HOME-OCT data ($10.89$). 
Overall, we can deduce that the proposed \unacornnb~has created a labeled data distribution with a similar uncertainty level to the unlabeled HOME-OCT data without directly learning the style or domain properties of the latter. 
\begin{table}[]
    \centering
    \begin{tabular}{c|c c c c}
    & real & synthetic&synthetic&synthetic\\
      real  & Spectralis  &  CycleGAN &  CUT& \unacornnb\\\hline
        HOME-OCT & 152.6&68.9&173.5&132.5 \\
         Spectralis & - &151.3&155.4&151.9\\
    \end{tabular}
    \caption{
    FID values for the created datasets compared to the real Spectralis and HOME-OCT data. 
    Small values indicate similar distributions.}
    \label{tab:FID}
\end{table}
\subsection{Application to Street Scenes}
\begin{figure}[tb]
    \centering
    \includegraphics[width=\linewidth]{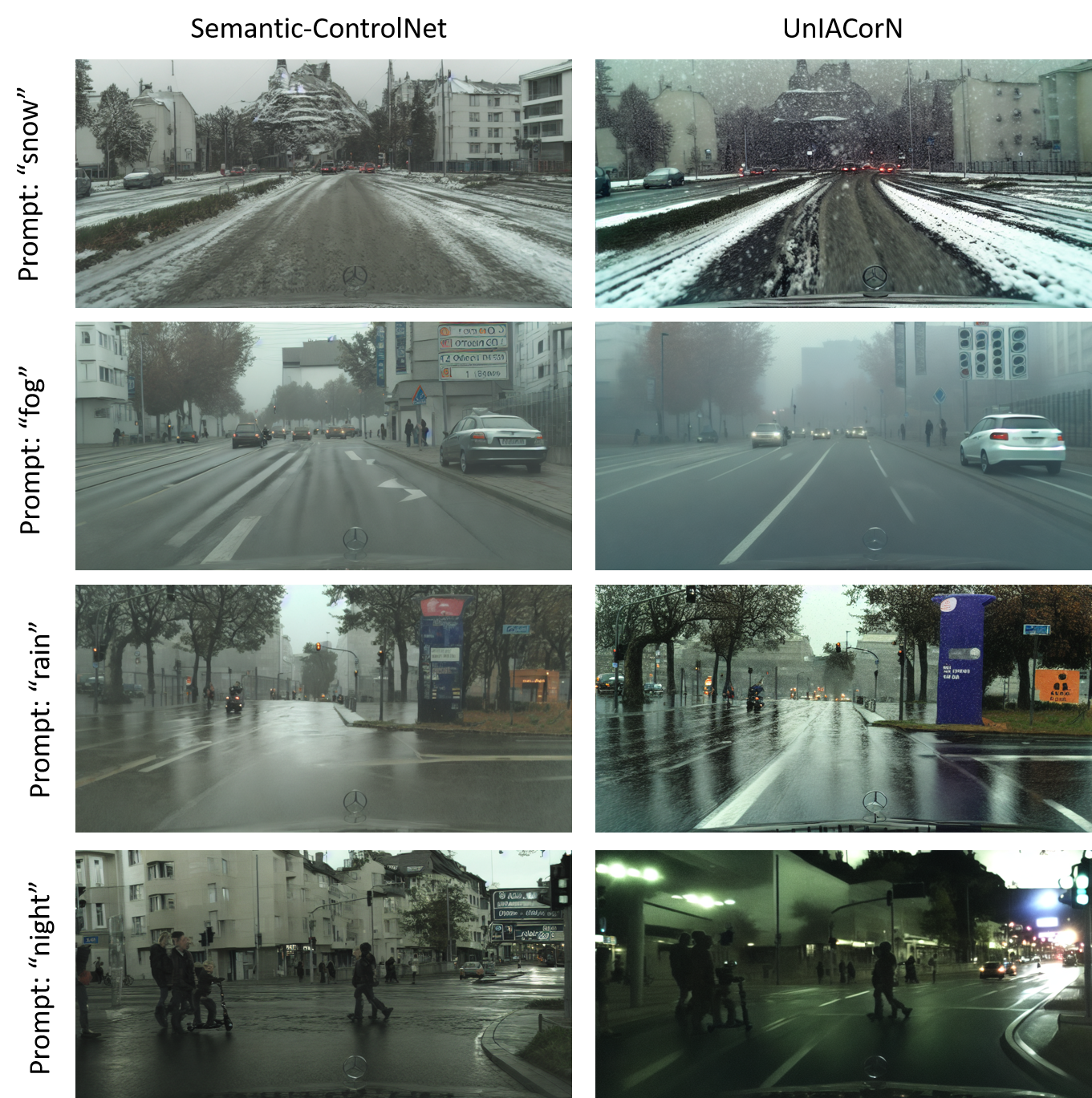} 
    \caption{The left column shows the Semantic-ControlNet result by a model (ALDM) trained on the Cityscapes dataset. 
    The right column shows the effect of the uncertainty guidance.
    Given the prompt, the image that is generated with \unacornnb\ is visibly more ``difficult'', making the segmentation network more uncertain (the fog scene gets foggier; the night scene becomes darker, etc.). }
    \label{fig:automotive}
\end{figure}

\begin{figure}[tb]
    \centering
    \includegraphics[width=\linewidth]{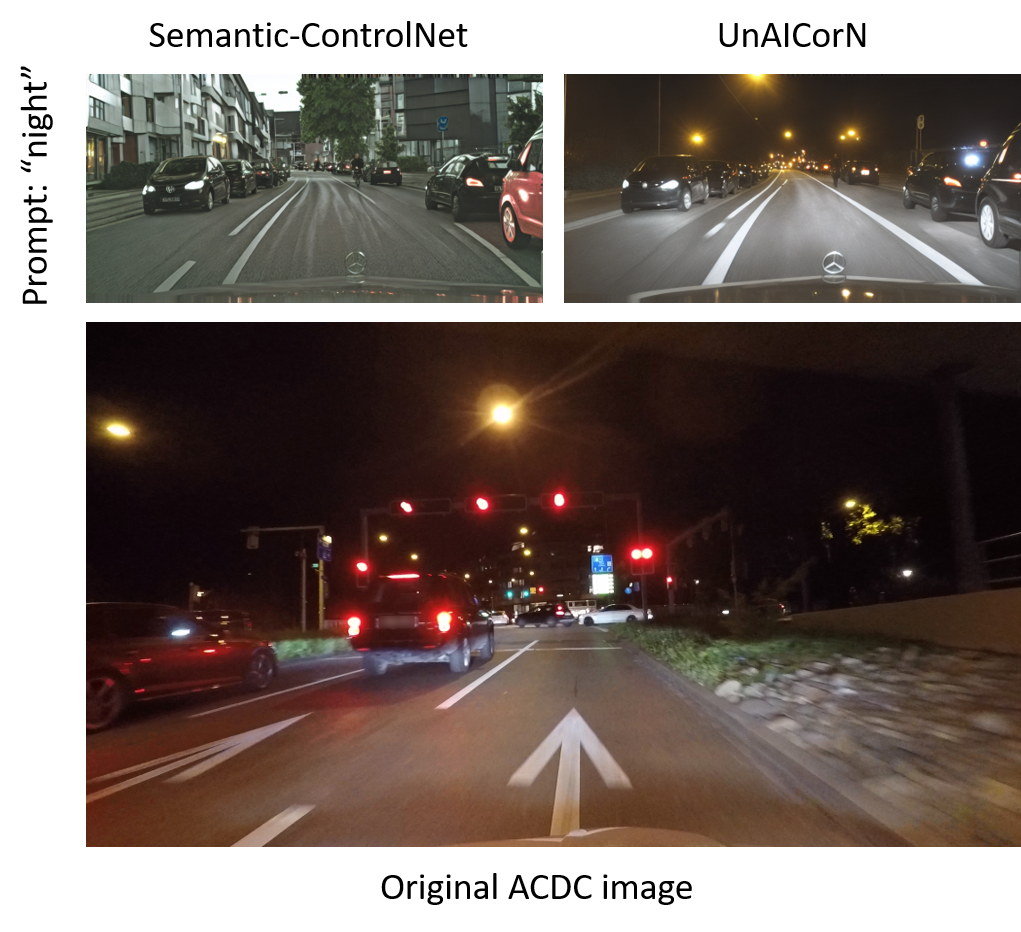}
    \caption{Comparison of an image generated by the baseline Semantic-ControlNet and our \unacornnb~approach with an original ACDC image of the subdomain from the prompt. 
    Since the \unacornnb\ has learned the correlation between uncertainty and image information on the ACDC dataset, the style matches better.}
    \label{fig:automotive_2}
\end{figure}

Since no explicit knowledge about the domains is integrated into \unacornnb, it becomes apparent that the proposed method is not yet another style transfer approach. 
\unacornnb's ability to generate data based on uncertainty enables the representation of arbitrary domain shifts without learning a particular image style, matching real-world scenarios where multiple (partly unknown) data domains are inhabiting a dataset. 
To show this generalization ability of our approach, we present a toy example on a street-scene use case, where our labeled distribution corresponds to the DCS data and the unlabeled distribution originates from the ACDC dataset. 
For this experiment, the backbone DDPM is a foundation model for RGB image generation~\cite{rombach2022high}, and the semantic ControlNet is based on the ALDM approach~\cite{li2024adversarial}. 
For the calculation of uncertainty, we utilize a SegFormer-based~\cite{DAFormer} architecture trained on the Cityscapes data. 
For data generation, we sample the uncertainty from $\mathcal{N}(9.6, 2.4)$ and generate an uncertainty-based dataset of size $2975$, which corresponds to generating a synthetic version of each Cityscapes train image (examples can be seen in \ref{fig:automotive}). 
An architectural difference of this approach is, however, the fact that the stable diffusion model additionally allows for influencing the generation process with text prompts (Fig.~\ref{fig:automotive}).
During the dataset creation, we include the four subdomains of the ACDC dataset (rain, fog, snow, and night) in the prompts, enabling targeted image generation for the baseline Semantic-ControlNet. 
The advantage of introducing uncertainty with our method, however, quickly becomes obvious. 
Since the control over the uncertainty in the image was trained on both the Cityscapes and the ACDC domain, the sub-domains of the unlabeled ACDC dataset, i.e., rain, snow, fog, or darkness, are generated more convincingly.  
This is because the Uncertainty-ControlNet learned the correlation of the uncertainties and the image information on the unlabeled data distribution.  
As Fig.~\ref{fig:automotive_2} shows, this causes the generated images of our \unacornnb~approach to be closer in style to the original ACDC data than the Semantic-ControlNet trained on Cityscapes. 
To evaluate this effect quantitatively, we used the sampled dataset to train the SegFormer-based model by Hoyer et al.~\cite{DAFormer} on the generated synthetic data simultaneously with the original Cityscapes data.
The resulting model was then evaluated on the ACDC domain. 
We achieved improvements of around $2\%$ in mIoU compared to the model only trained on the Cityscapes dataset and an improvement of around $1\%$ compared to the training on a dataset created with the same prompting strategy using the ALDM model.

\section{Discussion and Conclusion}
In this work, we proposed \unacornnb: a generative approach for synthesizing labeled image data with high epistemic uncertainty towards a downstream task. We propose a novel uncertainty-based conditioning mechanism for ControlNets, which enables the modeling of the uncertainties of a pre-trained discriminative model observed on recorded but unlabeled data. 
Such uncertainties could be interpreted as "known unknowns", since the image information that causes uncertain predictions is available; however, due to the lack of semantic labels, it cannot be utilized during the training process. 
In addition to uncertainty control, our method retains control over the semantic information, such as labels, during the generation process. In this way, both labeled and unlabeled data can be utilized during training, leading to an overall larger and more diverse training dataset and, eventually, to a more expressive generative model. 

This dual control strategy enables the sampling of diverse labeled training datasets based on uncertainties. As our experiments show, datasets generated by this approach lead to improved segmentation results as compared to standard style-transfer methods. 
Such methods are rigid in their learning of the explicit domain information, whereas the proposed method does not require any explicit knowledge about the domain (e.g., the type of acquisition device used for medical images) since it offers uncertainty-guidance based on out-of-distribution examples. 
Future work may explore this property more in depth by purposely presenting diverse domain shifts in the unlabeled dataset.

An underexplored advantage of the presented approach is its ability to sample multiple instances per input label map, unlike most style transfer models. 
For each label map, uncertainties can be sampled from a Gaussian distribution, corresponding to the distribution of the unlabeled data domain, and create various versions of the same label.
This property of the \unacornnb\ approach was not applied during the comparison with the baseline methods to avoid an unfair advantage.
In future work, however, we will explore this augmented generation, which is especially interesting for automotive use cases since car manufacturers usually have large amounts of unlabeled data from their fleets.
Another interesting future work, \unacornnb\, allows for the iterative improvement of the uncertainty modeling of the generative and discriminative models.
After training, the uncertainties of the discriminative model change, and hence, the generative model can be updated. This circular dependency opens doors for further advancements.

\section*{Acknowledgement}
The research leading to these results is funded by the German Federal Ministry for Economic Affairs and Energy within the project “NXT GEN AI METHODS – Generative Methoden für Perzeption, Prädiktion und Planung". The authors would like to thank the consortium for the successful cooperation.

{
    \small
    \bibliographystyle{ieeenat_fullname}
    \bibliography{main}
}

\end{document}